# Synthetic Similarity Search in Automotive Production

Christoph Huber[a,b,*], Ludwig Schleeh[b], Dino Knoll[b], Michael Guthe[a]

[a]*University of Bayreuth, Universitätsstraße 30, 95447 Bayreuth, Germany*
[b]*BMW Group, Petuelring 130, 80809 Munich, Germany*

\* Corresponding author. Tel.: +49-151-601-68945. *E-mail address:* christoph.ch.huber@bmw.de

**Abstract**

Visual quality inspection in automotive production is essential for ensuring the safety and reliability of vehicles. Computer vision (CV) has become a popular solution for these inspections due to its cost-effectiveness and reliability. However, CV models require large, annotated datasets, which are costly and time-consuming to collect. To reduce the need for extensive training data, we propose a novel image classification pipeline that combines similarity search using a vision-based foundation model with synthetic data. Our approach leverages a DINOv2 model to transform input images into feature vectors, which are then compared to pre-classified reference images using cosine distance measurements. By utilizing synthetic data instead of real images as references, our pipeline achieves high classification accuracy without relying on real data. We evaluate this approach in eight real-world inspection scenarios and demonstrate that it meets the high performance requirements of production environments.

## 1. Introduction

Visual quality control plays a critical role in the automotive production process, ensuring that vehicles meet stringent quality standards and customer expectations [4, 24]. It enables manufacturers to identify defects early in the production process, thereby preventing costly rework, delays, and recalls [19].

Today, many of these quality inspections are conducted using deep learning-based CV models, as they can deliver consistent performance, high accuracy, and the ability to analyze large volumes of data quickly [4, 11]. However, these models require extensive datasets during training to achieve the high performance levels necessary in a production setting [9, 13]. A major challenge in this context is that certain production variants may occur infrequently, making it difficult to collect a balanced dataset. Similarly, many types of defects are rare, which forces developers to either intentionally produce potentially costly defects or significantly extend the data collection period [10]. Additionally, changes to inspection stations or the introduction of new variants can render existing CV models ineffective, requiring new data collection and training efforts [14].

To address these challenges, many studies in the field of production have explored replacing real-world training images with synthetic data, as this approach enables the generation of large datasets with precise control over the content and distribution [1, 6, 9, 13]. However, the use of such datasets presents several challenges. To create high-quality images, developers must replicate complex materials, simulate realistic lighting, and accurately model camera characteristics, which demands significant time and resources [1]. Additionally, models trained with synthetic data typically suffer from the domain gap [20], which refers to the characteristic differences between real and synthetic images that lead to reduced model performance when applied to real-world inspection tasks.

In this paper, we present an alternative image classification pipeline for visual quality inspection stations in a vehicle assembly line. Inspections in this area generally concentrate on identifying the presence of relevant components and, in cases where various options are available due to national regulations or optional features, determining the exact variant. Frequent applications involve checking mounting hardware like screws and clamps, recognizing interior and exterior options, and confirming the presence of components that are easily missed.

Our image classification pipeline is based on the principle of similarity search [22]. Also known as nearest neighbor search, the goal is to identify the image from a labelled reference dataset that bears the closest resemblance to a given query image. We utilize a standard, task-agnostic DINOv2 model [17] to condense the relevant details from query and reference images into vectorized feature embeddings before employing k-Nearest Neighbor (k-NN) classification to find the most similar reference image and assign its label to the query image. In contrast to previous works [2, 16, 18], which require real images as references, we use available CAD data to generate synthetic reference images. This approach allows us to avoid the challenges associated with real-world data collection.

Through eight real-world inspection tasks, we demonstrate that this pipeline achieves high accuracy in image classification, making it a viable solution for real production environments. The key advantages of this approach include its independence from real data, the simplicity of generating synthetic images, and the absence of a need for task-specific model fine-tuning.

To the best of our knowledge, this is the first study to demonstrate that similarity search, combined with DINOv2 and synthetic data, can meet the high standards required for manufacturing applications. Our research opens the door for further exploration into a broader range of use cases. In the remainder of this paper, we will first discuss related work in the field. We will then proceed with a description of our image classification pipeline, followed by a demonstration of the use cases on which our research is based. Finally, we will present the details of our experiments and draw conclusions based on our findings.

## 2. Related Works

Researchers are exploring a multitude of methods to reduce the amount of real data needed for training AI models. One popular approach is the use of synthetic data. These studies utilize rendering engines such as Blender [26] to automatically generate massive amounts of artificial training images for a given task. There are two primary research domains within this area.

Some works aim to generate photo-realistic images that match the real world as closely as possible [1, 12, 14]. It has been shown that models trained with highly realistic images can generalize well between the synthetic and real domains. However, producing such images presents a significant challenge due to the complexity of accurately modeling surface textures and lighting conditions. This can quickly render photo-realistic data unprofitable, as the effort required to generate synthetic data can exceed that of collecting real images.

Other studies employ domain randomization [20] to produce unrealistic but highly randomized images. The underlying concept of domain randomization is that by introducing substantial variance into the training data, the trained model will perceive real-world data as just another variation within the same domain [7]. Many studies have demonstrated that models trained with this highly variant data can perform well on real images [6, 8, 13, 25]. However, rendering the required number of images is time-consuming, which can act as a bottleneck in fast-changing environments. Additionally, any changes to the inspection characteristics may necessitate re-training of the AI model.

In contrast, our pipeline utilizes similarity search, which identifies the item that is nearest to a query item, known as the nearest neighbor, from a reference database based on a distance measure [22]. In image processing, this typically involves using a convolutional neural network [18, 21] or a vision transformer [16, 23] to extract key features from an image, which are then compared to a set of references to find the most similar image using a k-NN algorithm. This paper specifically extends the work of Doerrich et al. [2], which utilizes DINOv2 [17] to extract image features. While these pipelines have demonstrated good performance on a variety of benchmark datasets, they all rely on real-world images for the reference dataset. However, in a production setting, it is not always feasible to collect sufficient data.

Our method, therefore, combines similarity search with synthetic data to achieve a pipeline that does not depend on real-world images, can operate with very few simple renderings as reference images, and does not require task-specific model fine-tuning.

## 3. Methodology

In this section, we discuss the concept of our image classification pipeline. Building on the work of Doerrich et al. [2], we combine similarity search with synthetic data to achieve an architecture that is independent of the availability of real-world data. The structure of the pipeline is illustrated in Fig. 1. It can be divided into three phases: reference image generation, knowledge extraction, and inference.

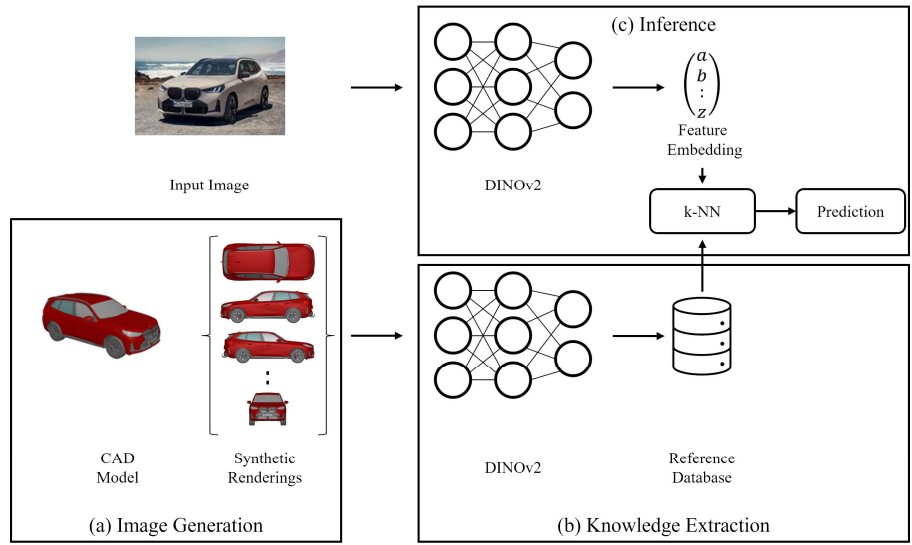

Fig. 1. The pipeline utilized in this paper for similarity search. During the image generation phase (a), we render simple synthetic images for each class using available CAD data. In the knowledge extraction phase (b), DINOv2 processes the reference images to extract the most relevant features, which are then stored as feature embeddings along with their corresponding labels in a database. Finally, during inference (c), the query image is processed by DINOv2, and the resulting feature embeddings are compared to the reference database using a k-NN algorithm with cosine similarity to identify the nearest neighbor.

*3.1. Reference Image Generation*

Similarity search requires a set of labeled reference images to compare against the query images. Ideally, these references should come from the same domain as the query data. However, especially in a manufacturing environment, it is not always feasible to collect a sufficient number of real images for all possible production variants and defects. Therefore, we utilize CAD data of relevant components to synthetically generate all necessary reference images. In most manufacturing companies, this data is readily available, since CAD is commonly used during the design and development of components [14].

For generating the images, we first import the CAD data into Blender, a versatile 3D creation suite utilized for modeling objects and generating renderings. We apply basic materials to all components in order to replicate the visual appearance of the actual images. A virtual camera is positioned within the scene in such a way that it captures the same perspective as the real images. During rendering, the camera's position and rotation are randomly adjusted around the initial orientation in such a way that the resulting images incorporate all possible perspective changes while ensuring that the region of interest remains visible at all times.

For each class, we render 24 images with a resolution of 224x224 pixels using the rendering engine 'EEVEE'.

*3.2. Knowledge Extraction*

In the knowledge-extraction phase, a pre-trained image encoder is employed to condense the most relevant features of the reference data into compact, vectorized representations. These feature embeddings are subsequently stored in a database along with the corresponding labels.

In this work, we use the DINOv2 [17] image encoder network for feature extraction. DINOv2 (self-**DI**stillation with **NO** labels) utilizes self-supervised learning to train a Vision Transformer [3] on a diverse set of 142M unlabeled images. It leverages a combination of image-level and patch-level objectives to learn robust visual features, which enables the model to effectively distinguish between different instances. DINOv2 is available in four different configurations, varying in the number of parameters and output feature size. The specifics for each configuration can be found in Table 1.

Table 1. Configurations of DINOv2 [17]. The inference times were measured on images with 224x224 pixels on a NVIDIA A6000 GPU.

| Configuration | # of params | Output Size | Inference Time |
|---|---|---|---|
| ViT-S/14 | 21 M | 384 | 8.2 ms |
| ViT-B/14 | 86 M | 768 | 9.6 ms |
| ViT-L/14 | 300 M | 1024 | 21 ms |
| ViT-g/14 | 1,100 M | 1536 | 58 ms |

*3.3. Inference*

During inference, a given query image is processed by DINOv2 to convert the image into feature embeddings. These embeddings are then compared to the entries in the reference database to identify the most similar items. We employ a k-NN algorithm with the cosine similarity metric to find the nearest neighbors of the query image. Cosine similarity measures the scale-invariant angular relationship between vectors, allowing it to effectively assess the similarity between multi-dimensional vectors [15]. The final classification of the query image is determined by a majority vote on the labels of the top-k most similar reference embeddings. In our experiments, we employ k = 5 for the k-NN classifier.

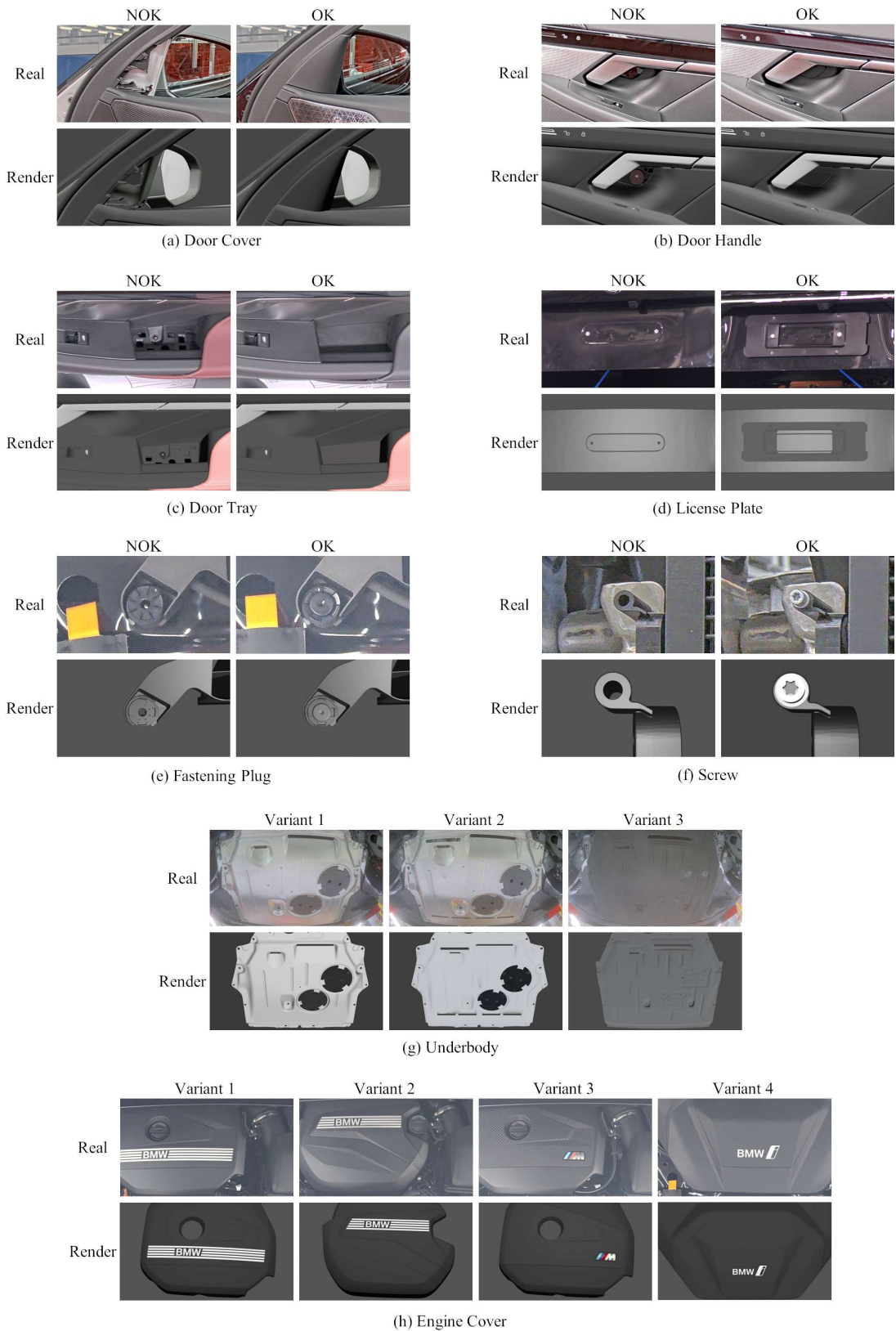

Fig. 2. Sample images from the eight use cases employed to verify the performance of our pipeline. Use cases (a) to (f) require the detection of the presence of a relevant object, while use cases (g) and (h) necessitate distinguishing between multiple variants of a component. The top row of each use case presents real images taken from the respective test set, while the bottom row displays synthetic renderings used as reference images for similarity search. These renderings feature simple materials and monochrome backgrounds.

## 4. Experiments

In this section, we discuss the experiments and results of our research, which demonstrate the advantages of this classification pipeline. We begin by introducing the use cases that form the basis of the study. Following this, we present the performance of the pipeline on these use cases. We then analyze the influence of the DINOv2 configuration used on classification accuracy.

### 4.1. Test Cases

We evaluate the applicability of our pipeline to visual quality inspection across eight real-world inspection use cases, as presented in Fig. 2.

These inspections can be divided into two categories: inspections (a) to (f) determine the presence or absence of relevant components (e.g., screws, covers), while inspections (g) and (h) differentiate between various variants of the same component. These examples are representative of numerous inspections commonly encountered in manufacturing plants.

We assess the performance of our pipeline on these use cases using test sets that contain real images taken from the production line. Each test set contains 200 images per class. The images are resized to 224x224 pixels before inference.

### 4.2. Experimental Evaluation

In this section, we evaluate the classification performance of our proposed pipeline, which combines similarity search using DINOv2 with synthetic data. The evaluation focuses on the eight inspection use cases presented in Section 4.1. Table 2 displays the F1-scores achieved by our pipeline for the respective use cases. The results are differentiated according to the variant of DINOv2 used for the image embeddings.

Table 2. Classification performance of our pipeline for different choices of DINOv2 configurations. We report the F1-scores achieved on the use cases presented in Fig. 2.

| Use Case | ViT-S/14 | ViT-B/14 | ViT-L/14 | ViT-g/14 |
|---|---|---|---|---|
| Door Cover | 1.00 | 1.00 | 0.98 | 0.98 |
| Door Handle | 1.00 | 0.98 | 0.96 | 1.00 |
| Door Tray | 1.00 | 1.00 | 1.00 | 1.00 |
| License Plate | 1.00 | 0.99 | 0.95 | 0.96 |
| Fastening Plug | 1.00 | 1.00 | 1.00 | 1.00 |
| Underbody | 1.00 | 0.91 | 0.78 | 0.85 |
| Screw | 0.96 | 0.99 | 0.99 | 0.99 |
| Engine Cover | 1.00 | 1.00 | 1.00 | 0.99 |

These performance values show that our pipeline is able to achieve high classification accuracy without the need for real data. When comparing the individual DINOv2 models against each other, it becomes evident that ViT-S/14 achieves particularly high accuracies, with perfect results on seven out of the eight test cases, only falling behind the other models on one occasion. These findings contradict the results of Doerrich et al. [2], who found that larger DINOv2 models outperformed smaller ones in their tests. This discrepancy may be attributed to the fact that DINOv2 was primarily trained on natural images [17]. We hypothesize that the characteristic differences between DINOv2's training data and our inspection images are too significant, leading to larger models struggling to extract meaningful features from the images. In contrast, the smaller size of ViT-S/14 may have acted as a form of regularization, compelling the model to focus on more fundamental image features, which proves beneficial in our application.

## 5. Discussion

In this section, we address some potential limitations of our study. The first concern is our selection of eight use cases, each with only a small number of classes to be distinguished, to demonstrate the applicability of our classification pipeline for automotive visual quality inspection. While this may seem limited, this selection represents a significant number of use cases typically encountered in the field.

The second limitation is that this pipeline may struggle to achieve good performance on inspections that are more complex than the presented use cases. During our experiments, it became apparent that DINOv2 was unable to extract meaningful features from images that are poorly lit, heavily noisy, or blurred, leading to poor classification accuracy. Furthermore, in cases where multiple possible prediction classes exhibit very similar geometries - such as various types of screws with marginally different sizes - the extracted feature embeddings were too close together, thus prohibiting reliable classification. Nevertheless, even in such situations, this pipeline can still serve as a cost-efficient and easy-to-implement interim solution during the data collection phase for a task-specific AI model.

One possible enhancement to improve the performance of the pipeline in such challenging applications would be domain-specific fine-tuning of DINOv2 specifically for automotive visual quality inspection. By training DINOv2 on a dataset containing images from a wide variety of current inspection use cases, it is possible that the resulting foundation model will perform better on future inspections than the standard version of DINOv2. However, this is beyond the scope of this paper and should be addressed in future work.

## 6. Conclusion

In this work, we present an image classification pipeline that combines similarity search using DINOv2 with synthetic data for visual quality inspection in automotive production. Instead of relying on real images as references for classification, we use simple synthetic images rendered from available CAD data. Through eight real-world inspection scenarios, we demonstrate that this approach enables the development of a pipeline that achieves high classification accuracy without requiring any real images. We would like to emphasize the key advantages of this approach. By leveraging both similarity search and synthetic data, our pipeline enables the creation of high-quality image classifiers while significantly reducing development time and costs. Furthermore, its flexible design makes it well-suited for industrial applications, since it allows for the seamless integration of environmental changes and new parts or defect types without the need for model retraining. Future work should explore the applicability of this pipeline across a broader range of use cases that involve, for example, different industries and challenging environmental influences.

**References**

[1] C. Abou-Akar et al., "Synthetic Object Recognition Dataset for Industries". In 2022 35th SIBGRAPI Conference on Graphics, Patterns and Images (SIBGRAPI), 150–55. Natal, Brazil: IEEE, 2022.

[2] S. Doerrich et al., "Integrating kNN with Foundation Models for Adaptable and Privacy-Aware Image Classification". In 2024 IEEE International Symposium on Biomedical Imaging (ISBI), 1–5, 2024.

[3] A. Dosovitskiy et al., "An Image is Worth 16x16 Words: Transformers for Image Recognition at Scale". arXiv, 3. June 2021.

[4] A. Ebayyeh, A. Mousavi, "A Review and Analysis of Automatic Optical Inspection and Quality Monitoring Methods in Electronics Industry". IEEE Access 8 (2020): 183192–271.

[5] L. Eversberg, J. Lambrecht, "Generating Images with Physics-Based Rendering for an Industrial Object Detection Task: Realism versus Domain Randomization". Sensors 21, Nr. 23 (January 2021): 7901.

[6] I. Gräßler, M. Hieb, "Creating Synthetic Datasets for Deep Learning used in Machine Vision". Procedia CIRP, 17th CIRP Conference on Intelligent Computation in Manufacturing Engineering (CIRP ICME '23), 126 (1. January 2024): 981–86.

[7] S. Hinterstoisser et al., "An Annotation Saved is an Annotation Earned: Using Fully Synthetic Training for Object Detection". In 2019 IEEE/CVF International Conference on Computer Vision Workshop (ICCVW), 2787–96. Seoul, Korea (South): IEEE, 2019.

[8] D. Horváth et al., "Object Detection Using Sim2Real Domain Randomization for Robotic Applications". IEEE Transactions on Robotics 39, Nr. 2 (April 2023): 1225–43.

[9] C. Huber et al., "Fully-Synthetic Training for Visual Quality Inspection in Automotive Production". arXiv, 12. March 2025.

[10] S. Jain et al., "Synthetic Data Augmentation for Surface Defect Detection and Classification Using Deep Learning". Journal of Intelligent Manufacturing 33, Nr. 4 (April 2022): 1007–20.

[11] F. Konstantinidis et al., "The Role of Machine Vision in Industry 4.0: an automotive manufacturing perspective". In 2021 IEEE International Conference on Imaging Systems and Techniques (IST), 1–6. Kaohsiung, Taiwan: IEEE, 2021.

[12] C. Manettas et al., "Synthetic datasets for Deep Learning in computer-vision assisted tasks in manufacturing". Procedia CIRP, 9th CIRP Global Web Conference – Sustainable, resilient, and agile manufacturing and service operations: Lessons from COVID-19, 103 (1. January 2021): 237–42.

[13] C. Mayershofer et al., "Towards Fully-Synthetic Training for Industrial Applications". In 10th International Conference on Logistics, Informatics and Service Sciences (LISS), 765–82. Singapore: Springer Singapore, 2021.

[14] S. Moonen et al., "CAD2Render: A Modular Toolkit for GPU-Accelerated Photorealistic Synthetic Data Generation for the Manufacturing Industry". In 2023 IEEE/CVF Winter Conference on Applications of Computer Vision Workshops (WACVW), 583–92. Waikoloa, HI, USA: IEEE, 2023.

[15] M. S. Mulekar et al., "Distance and Similarity Measures". In Encyclopedia of Social Network Analysis and Mining, 1–16. New York, NY: Springer, 2017.

[16] K. Nakata et al., "Revisiting a kNN-Based Image Classification System with High-Capacity Storage". In Computer Vision – ECCV 2022, 457–74. Cham: Springer Nature Switzerland, 2022.

[17] M. Oquab et al., "DINOv2: Learning Robust Visual Features without Supervision". arXiv, 2. Februar 2024.

[18] S. Sani et al., "Learning deep features for kNN-based human activity recognition". 95–103. CEUR-WS, 2017.

[19] J. See, "Visual Inspection: A Review of the Literature." Sandia National Laboratories (SNL), Albuquerque, NM, and Livermore, CA (United States), 1. October 2012.

[20] J. Tobin et al., "Domain randomization for transferring deep neural networks from simulation to the real world". In 2017 IEEE/RSJ International Conference on Intelligent Robots and Systems (IROS), 23–30. Vancouver, BC: IEEE, 2017.

[21] J. T. Turner et al., "Novel Object Discovery Using Case-Based Reasoning and Convolutional Neural Networks". In Case-Based Reasoning Research and Development, 399–414. Cham: Springer International Publishing.

[22] J. Wang et al., "Hashing for Similarity Search: A Survey". arXiv, 13. August 2014.

[23] Z. Zhao et al., "Case-Enhanced Vision Transformer: Improving Explanations of Image Similarity with a ViT-based Similarity Metric". arXiv, 24. July 2024.

[24] L. Zhou et al., "Computer Vision Techniques in Manufacturing". IEEE Transactions on Systems, Man, and Cybernetics: Systems 53, Nr. 1 (January 2023): 105–17.

[25] X. Zhu et al., "Towards Sim-to-Real Industrial Parts Classification with Synthetic Dataset". In 2023 IEEE/CVF Conference on Computer Vision and Pattern Recognition Workshops (CVPRW), 4454–63. Vancouver, BC, Canada: IEEE, 2023.

[26] Blender, The Freedom to Create, Available: https://www.blender.org/download/.